\documentclass{article}

\usepackage{microtype}
\usepackage{graphicx}
\usepackage{subcaption}
\usepackage{booktabs} % for professional tables
\usepackage{hyperref}
\usepackage{amsmath, amssymb}
\usepackage{xcolor}
\usepackage{url}

\usepackage{hyperref}

\usepackage[accepted]{icml2026}

\usepackage{amsmath}
\usepackage{amssymb}
\usepackage{mathtools}
\usepackage{amsthm}

\usepackage[capitalize,noabbrev]{cleveref}

\theoremstyle{plain}

\theoremstyle{definition}

\theoremstyle{remark}

\usepackage[textsize=tiny]{todonotes}

\icmltitlerunning{Selective Memory Retention for Long-Horizon LLM Agents}

\begin{document}

\twocolumn[
  \icmltitle{Selective Memory Retention for Long-Horizon LLM Agents}

  % It is OKAY to include author information, even for blind submissions: the
  % style file will automatically remove it for you unless you've provided
  % the [accepted] option to the icml2026 package.

  % List of affiliations: The first argument should be a (short) identifier you
  % will use later to specify author affiliations Academic affiliations
  % should list Department, University, City, Region, Country Industry
  % affiliations should list Company, City, Region, Country

  % You can specify symbols, otherwise they are numbered in order. Ideally, you
  % should not use this facility. Affiliations will be numbered in order of
  % appearance and this is the preferred way.
  \icmlsetsymbol{equal}{*}

  \begin{icmlauthorlist}
    \icmlauthor{Pranath Reddy}{indep}
    % \icmlauthor{Firstname1 Lastname1}{equal,yyy}
    % \icmlauthor{Firstname2 Lastname2}{equal,yyy,comp}
    % \icmlauthor{Firstname3 Lastname3}{comp}
    % \icmlauthor{Firstname4 Lastname4}{sch}
    % \icmlauthor{Firstname5 Lastname5}{yyy}
    % \icmlauthor{Firstname6 Lastname6}{sch,yyy,comp}
    % \icmlauthor{Firstname7 Lastname7}{comp}
    %\icmlauthor{}{sch}
    %\icmlauthor{Firstname8 Lastname8}{sch}
    %\icmlauthor{Firstname8 Lastname8}{yyy,comp}
    %\icmlauthor{}{sch}
    %\icmlauthor{}{sch}
  \end{icmlauthorlist}

  \icmlaffiliation{indep}{Independent Researcher, Huntsville, Alabama, USA}
  % \icmlaffiliation{yyy}{Department of XXX, University of YYY, Location, Country}
  % \icmlaffiliation{comp}{Company Name, Location, Country}
  % \icmlaffiliation{sch}{School of ZZZ, Institute of WWW, Location, Country}

  \icmlcorrespondingauthor{Pranath Reddy}{kumbam.pranath@gmail.com} 
  %\icmlcorrespondingauthor{Firstname1 Lastname1}{first1.last1@xxx.edu}
  %\icmlcorrespondingauthor{Firstname2 Lastname2}{first2.last2@www.uk}

  % You may provide any keywords that you find helpful for describing your
  % paper; these are used to populate the "keywords" metadata in the PDF but
  % will not be shown in the document
  \icmlkeywords{LLM agents, episodic memory, retention, continual adaptation}

  \vskip 0.3in
]

% this must go after the closing bracket ] following \twocolumn[ ...

% This command actually creates the footnote in the first column listing the
% affiliations and the copyright notice. The command takes one argument, which
% is text to display at the start of the footnote. The \icmlEqualContribution
% command is standard text for equal contribution. Remove it (just {}) if you
% do not need this facility.

% Use ONE of the following lines. DO NOT remove the command.
% If you have no special notice, KEEP empty braces:
\printAffiliationsAndNotice{}  % no special notice (required even if empty)
% Or, if applicable, use the standard equal contribution text:
% \printAffiliationsAndNotice{\icmlEqualContribution}

\begin{abstract}
When does retention matter for memory-augmented LLM agents? We study this with \textbf{TraceRetain}, a lightweight framework for bounded external memory in frozen LLM agents that scores entries by interpretable features (success, age, access frequency, redundancy, specificity, similarity, downstream utility) and evicts the lowest-scoring ones at capacity. On clean ALFWorld with \texttt{gpt-5-mini}, external memory robustly improves over no memory across two seeds, but differences among bounded retention policies fall within Wilson 95\% CIs: clean ALFWorld at $T{=}100$ to $T{=}200$ does not naturally exhibit the memory pollution retention is designed to address. Under a controlled noisy-write stress (75\% synthetic distractors), unbounded memory and FIFO-K50 degrade on Precision@5 (20.2\%$\to$12.4\% and 15.8\%$\to$3.8\%) while TraceRetain-CEM is essentially unchanged (16.9\%$\to$16.6\%) and preserves 97/100 task success. The mechanism: unbounded memory has the highest mean similarity (0.87) but lowest precision, indicating failed distractors close to the query in embedding space. Held-out in-distribution evaluation shows memory-augmented policies solving 47--49 of 50 tasks vs.\ 39/50 for no memory. Bounded retention buys memory and step efficiency on saturated clean benchmarks at no task-success cost, and only differentiates from cache heuristics when streams contain noise.
\end{abstract}
 
\section{Introduction}
\label{sec:intro}
 
LLM agents increasingly rely on external memory to act over long horizons \citep{shinn2023reflexion,wang2023voyager,park2023generative,packer2024memgpt}. Storing successful trajectories and retrieving relevant prior experience supports non-parametric adaptation: behavior changes as the memory bank changes, without updating the base model.
 
Yet more memory is not automatically better. As an agent stores more episodes, retrieval must search through a growing set of similar but not necessarily useful experiences. Redundant memories can crowd out rare but important ones, failed trajectories can be retrieved for new tasks, and generic examples can consume prompt budget without helping the agent choose better actions. We call this risk \textbf{memory pollution}: degradation in retrieval quality or downstream behavior caused by unbounded or poorly managed accumulation.
 
This paper asks a deliberately empirical question: in current memory-augmented agent benchmarks, when does retention actually matter? The answer has direct implications for sustainable continual adaptation. Bounded memory caps both the candidate set scored at retrieval time and the prompt token footprint over long task streams, so a positive answer translates into inference cost reduction; a negative answer means engineers can default to simple cache heuristics without loss. Either outcome is informative for resource-aware deployment \citep{strubell2019energy,patterson2021carbon}.
 
We study this question with \textbf{TraceRetain}, a lightweight retention framework that freezes the base LLM and scores memories with a small linear function over interpretable features. When memory exceeds capacity, the lowest-scoring entries are evicted. We use TraceRetain as a probe rather than as a winning method, comparing it against cache heuristics (FIFO, LRU, LFU, Random, Ebbinghaus decay) and against unbounded memory.
 
Our findings are characterizations rather than method rankings. With a strong base agent (\texttt{gpt-5-mini}), clean ALFWorld \citep{shridhar2021alfworld} at $T{=}100$ and $T{=}200$ is highly saturated: most bounded retention policies cluster within Wilson 95\% confidence intervals of unbounded memory, and pollution does not appear naturally at these horizons. When we introduce a controlled noisy-write stress, unbounded memory does degrade and bounded retention helps; in this regime TraceRetain-CEM separates from cache heuristics on retrieval precision while preserving task success at half the memory. Retrieval diagnostics show that pollution manifests as high embedding similarity to failed distractors rather than as low similarity, the failure mode insertion-order eviction is blind to.
 
\paragraph{Contributions.}
\begin{enumerate}
\item We formulate external-memory management as a capacity-constrained retention problem and connect bounded retention to inference cost over long task streams.
\item We introduce TraceRetain, a lightweight retention framework with interpretable features and two scorer variants: TraceRetain-Linear (fixed weights) and TraceRetain-CEM (cross-entropy weight search).
\item We report Wilson 95\% CIs and exact paired sign tests on all conditions, and identify the regime where bounded retention separates from cache heuristics: noisy-write streams and held-out in-distribution evaluation, not clean saturated benchmarks.
\item We provide a controlled diagnostic that produces memory pollution on demand, and surface a retrieval-level mechanism (high similarity, low precision) that explains why insertion-order eviction fails under pollution while task-aware retention does not.
\end{enumerate}
 
\section{Related Work}
\label{sec:related}
 
\textbf{Memory-augmented LLM agents.} Frozen-LLM agents are routinely augmented with external state for non-parametric adaptation: Reflexion stores verbal self-critiques after failures \citep{shinn2023reflexion}, Voyager grows a skill library for open-ended exploration \citep{wang2023voyager}, ExpeL distills trajectory insights \citep{zhao2024expel}, MemGPT manages context hierarchically \citep{packer2024memgpt}, MemoryBank applies psychologically inspired decay \citep{zhong2024memorybank}, and Generative Agents organize observations as time-stamped streams with importance scores \citep{park2023generative}. These systems concentrate primarily on \emph{what to write} and \emph{what to retrieve}; capacity-bounded eviction under long task streams is comparatively under-studied. Cognitive-architecture surveys map the broader design space \citep{sumers2024cognitive}. We treat retention as the explicit object of study and benchmark learned scorers against simple cache heuristics under matched capacity.
 
\textbf{Episodic memory and retrieval-augmented reasoning.} Bounded experience replay is well established in continual learning, where reservoir sampling, gradient episodic memory, and rehearsal buffers manage capacity under stream pressure \citep{lopezpaz2017gem,rolnick2019experience}. We adapt this framing to the LLM-agent setting, where ``replay'' is in-context retrieval rather than gradient update, so eviction interacts with embedding-based retrieval rather than parametric forgetting. RAG \citep{lewis2020rag} and reasoning-action paradigms \citep{yao2023react,schick2023toolformer} extend retrieval to interactive settings, but their indices are typically curated corpora; ours grows from the agent's own trajectories during the task stream, so retention quality is endogenous to the policy. ALFWorld \citep{shridhar2021alfworld} extends TextWorld \citep{cote2018textworld} to embodied household tasks, and we use it because its task families and outcome labels make pollution and retrieval relevance directly measurable; WebShop \citep{yao2022webshop} is a complementary multi-turn benchmark.
 
\section{Method}
\label{sec:method}
 
\paragraph{Episodic memory setting.}
An agent solves a sequence of tasks indexed by episode $t$. At each step it embeds the task, retrieves up to $k$ memories from a bank $\mathcal{M}_t$, and acts using a frozen ReAct-style LLM \citep{yao2023react}. After the episode, a trajectory summary (task, outcome, interaction history) is added to $\mathcal{M}_t$. The bank has capacity $K$; if $|\mathcal{M}_t| > K$, the retention policy must evict.
 
\paragraph{Features.}
For each memory $m_i$ at episode $t$, we first compute a feature vector
$\phi(m_i,t)$ whose components include success/failure, normalized age,
last-access gap, log access frequency, specificity, redundancy, step
efficiency, observed downstream utility, utility-count confidence, last
retrieval similarity, and average retrieval similarity. The retention
score is $s_i = w^\top \phi(m_i,t)$, where $w$ is a vector of feature
weights. When $|M_t| > K$, the policy evicts the lowest-scoring entries.
Appendix~\ref{app:task-memory-example} gives a concrete task and memory
example.
 
\paragraph{Scorer variants.}
\textbf{TraceRetain-Linear} uses fixed feature weights chosen to reward successful, specific, frequently accessed memories and penalize old, unused, redundant, or failed ones; it requires no extra LLM calls. \textbf{TraceRetain-CEM} uses cross-entropy-method search \citep{rubinstein1999cem,deboer2005cem} over the same feature weights, with candidate scorers evaluated on a 20-task tuning subset and ranked by a scalar combining success, step efficiency, and retrieval precision. Because the tuning subset overlaps the seed-42 train stream, the seed-43 and eval-seen results are the cleaner generalization checks. We name this variant CEM rather than RL because the optimization is black-box population search, not online reinforcement learning.
 
\paragraph{Noisy-write stress.}
To probe pollution directly, we add a condition in which each real memory write is followed by three failed same-task distractor entries, so the bank is 75\% synthetic noise by construction. Distractors share the current task description and embedding but contain failed trajectory content. This is a diagnostic stress, not a model of natural ALFWorld noise.
 
\paragraph{Baselines.}
We compare against no memory, unbounded memory, FIFO-K50 (first-in-first-out), LRU-K50 (least recently used), LFU-K50 (least frequently used), Random-K50, Ebbinghaus-K50 (decay-based), and an exploratory offline policy-gradient variant TraceRetain-RL-K50. We report the strongest in the main tables and full results in the appendix.
 
\begin{table*}[t]
\centering
\small
\caption{Success counts across conditions. Clean $T{=}100$ column is the mean of two seeds (42, 43); its bracketed CI is a pooled Wilson 95\% CI computed over the combined $n{=}200$ task pool. TraceRetain-CEM weights were tuned on a seed-42 train subset; seed-43 and eval-seen are unbiased generalization checks. Bracketed values for single-seed cells are Wilson 95\% CIs in percentage points.}
\label{tab:headline}
\begin{tabular}{lcccc}
\toprule
Policy & Clean $T{=}100$ (mean of 2 seeds) & Clean $T{=}200$ & Noisy $T{=}100$ & Eval (50) \\
\midrule
No memory          & 86.0/100\;[80,90] & 182/200\;[86,94] & 88/100\;[80,93] & 39/50\;[65,87] \\
Unbounded          & 97.5/100\;[94,99] & 192/200\;[92,98] & 95/100\;[89,98] & 47/50\;[84,98] \\
FIFO-K50           & 96.0/100\;[92,98] & 190/200\;[91,97] & 94/100\;[88,97] & 48/50\;[87,99] \\
TraceRetain-Linear-K50 & 97.5/100\;[94,99] & 190/200\;[91,97] & 96/100\;[90,98] & 48/50\;[87,99] \\
TraceRetain-CEM-K50    & 96.0/100\;[92,98] & 192/200\;[92,98] & \textbf{97/100}\;[92,99] & \textbf{49/50}\;[90,100] \\
\bottomrule
\end{tabular}
\end{table*}
 
\section{Experiments}
\label{sec:exp}
 
We evaluate on ALFWorld household manipulation tasks \citep{shridhar2021alfworld} with \texttt{gpt-5-mini}, \texttt{text-embedding-3-large}, top-5 retrieval, max 50 environment steps, and $K{=}50$ unless specified. We report five conditions: clean $T{=}100$ with seed 42 and seed 43, clean $T{=}200$ with seed 42, noisy-write $T{=}100$ with seed 42, and a transfer condition that writes memory on 100 train tasks and then evaluates on 50 held-out \texttt{eval\_in\_distribution} tasks without further writes. The agent uses a frozen ReAct-style prompt with a fixed task-specific few-shot example. Memories are retrieved by embedding similarity. Precision@5 uses normalized ALFWorld task keys: a retrieved memory is relevant if it is successful and either matches the task key exactly or shares the task family with the same target object or receptacle (see Appendix~\ref{app:precision}). We treat Precision@5 as a diagnostic proxy; task success is the primary metric. The full experimental footprint is approximately 4000 ALFWorld episodes with no parameter updates at any point.
 
\paragraph{Statistical reporting.}
For sample sizes used here ($n{=}50$ to $n{=}200$ per condition), Wilson 95\% CI half-widths near the saturation ceiling are roughly $\pm 3$ to $\pm 8$ percentage points; we report exact CIs and treat overlapping CIs as inconclusive. For memory-vs-no-memory claims we additionally report exact paired sign tests over the same task sequence, which control for per-task difficulty and are stronger than marginal CI comparisons.
 
\section{Results}
\label{sec:results}
 
\subsection{Clean ALFWorld: Memory Helps, Methods Saturate}
\label{sec:results-clean}
 
Across two seeds at $T{=}100$, all memory policies outperform no memory by 9 to 12 tasks on average. Paired sign tests on the seed-42 stream are significant for every memory policy against no memory (e.g., TraceRetain-Linear gains 11 tasks and loses 0, $p<0.001$; FIFO-K50 gains 9 and loses 0, $p{=}0.004$). Differences among bounded methods are small and within Wilson 95\% CIs at single-seed resolution. Table~\ref{tab:headline} reports the consolidated picture across all four conditions.
 
On clean $T{=}100$, TraceRetain-Linear ties unbounded memory on the two-seed mean (97.5/100) while using half the memory. On the longer $T{=}200$ run, TraceRetain-CEM ties unbounded (192/200) at one quarter of the final memory size. We do not claim that learned retention robustly dominates cache heuristics on clean ALFWorld: at $T{=}200$, FIFO-K50 ties TraceRetain-Linear at 190/200, and on seed 43 Random-K50 ties TraceRetain-Linear at 96/100. The defensible clean-run claim is memory efficiency under fixed capacity, not method ranking among bounded policies.
 
\subsection{Noisy-Write Stress: A Controlled Pollution Regime}
\label{sec:results-noisy}
 
When 75\% of writes are failed same-task distractors, the unbounded bank grows to 400 entries by episode 100 and methods separate clearly (Table~\ref{tab:noisy}).
 
\begin{table}[t]
\centering
\footnotesize
\caption{Noisy-write $T{=}100$, seed 42. Bracketed values are Wilson 95\% CIs. TR- prefixes denote TraceRetain variants.}
\label{tab:noisy}
\setlength{\tabcolsep}{4pt}
\begin{tabular}{lcccc}
\toprule
Policy & Mem & Success [95\% CI] & Steps & P@5 \\
\midrule
TR-CEM-K50    & 50  & \textbf{97/100}\;[92,99] & 11.61 & \textbf{16.6\%} \\
TR-Linear-K50 & 50  & 96/100\;[90,98] & 12.34 & 15.4\% \\
Unbounded          & 400 & 95/100\;[89,98] & 12.57 & 12.4\% \\
FIFO-K50           & 50  & 94/100\;[88,97] & 12.63 & 3.8\% \\
No memory          & 0   & 88/100\;[80,93] & 18.35 & 0.0\% \\
\bottomrule
\end{tabular}
\end{table}
 
The success-rate Wilson CIs across bounded methods overlap, so the strongest claim from this table is not the success ordering. The discriminating signals are Precision@5 stability and a paired-sign-test asymmetry against no memory. Across the clean$\rightarrow$noisy transition, Unbounded falls from 20.2\% to 12.4\% Precision@5, FIFO collapses from 15.8\% to 3.8\%, while TraceRetain-CEM is essentially unchanged (16.9\% to 16.6\%) and TraceRetain-Linear loses less than one percentage point (16.1\% to 15.4\%). Under paired sign tests against no memory in this regime, TraceRetain-CEM (10 gained, 1 lost, $p{=}0.012$) and TraceRetain-Linear ($p{=}0.021$) reach significance, while Unbounded ($p{=}0.065$) and FIFO ($p{=}0.109$) do not: when noise is present, only retention-aware methods reliably improve over the no-memory baseline.
 
Retrieval diagnostics surface the mechanism. Unbounded achieves the highest mean retrieval similarity (0.87) but only 12.4\% Precision@5: the polluted bank places failed distractors close to the query in embedding space, so similarity-based retrieval surfaces them. TraceRetain-CEM trades slightly lower similarity (0.82) for higher precision (16.6\%) by discounting failed entries via the success and redundancy features. FIFO has comparable similarity (0.81) to TraceRetain-CEM but collapses to 3.8\% precision because insertion-order eviction is blind to pollution.
 
This regime is synthetic and was constructed specifically to stress retention. We do not claim that ordinary ALFWorld writes contain this level of noise. The result is best read as a controlled diagnostic: the proposed method discriminates cleanly when pollution is present, and FIFO does not. Whether and when natural agent streams produce comparable pollution is the open empirical question this characterization is meant to surface.
 
\subsection{Held-Out Eval-Seen Transfer}
\label{sec:results-eval}
 
We next test whether memory written during 100 train tasks transfers to held-out tasks without further writes (Table~\ref{tab:eval}).
 
\begin{table}[t]
\centering
\footnotesize
\caption{Eval-seen transfer. Memory is written during the 100 train tasks; the 50 eval tasks use retrieval but do not modify memory. Bracketed values are Wilson 95\% CIs.}
\label{tab:eval}
\setlength{\tabcolsep}{3pt}
\begin{tabular}{lccccc}
\toprule
Policy & Train & Eval [95\% CI] & Steps & P@5 & Mem \\
\midrule
TR-CEM-K50    & 96/100 & \textbf{49/50}\;[90,100] & 9.52  & 29.2\% & 50  \\
TR-Linear-K50 & 97/100 & 48/50\;[87,99]  & 12.10 & 30.0\% & 50  \\
FIFO-K50           & 96/100 & 48/50\;[87,99]  & 11.08 & 26.4\% & 50  \\
Unbounded          & 96/100 & 47/50\;[84,98]  & 10.82 & 45.2\% & 100 \\
No memory          & 87/100 & 39/50\;[65,87]  & 20.98 & 0.0\%  & 0   \\
\bottomrule
\end{tabular}
\end{table}
 
The dominant signal is again memory versus no memory: every memory-augmented policy solves 47 to 49 eval tasks (Wilson 95\% CIs from $\approx[0.84, 1.00]$) and beats no memory under paired sign tests at $p<0.001$, while no memory solves 39 ($\approx[0.65, 0.87]$). TraceRetain-CEM also reaches the lowest eval step count (9.52) of any policy, beating even unbounded memory (10.82) at half the storage. Differences among bounded methods are at most one to two tasks with heavily overlapping CIs at $n{=}50$ and we do not over-read them. The transfer result is sufficient to address the train-only validity concern: bounded retention generalizes to held-out in-distribution tasks at least as well as unbounded memory while storing fewer entries. Unbounded memory has the highest eval Precision@5 (45.2\%) but the lowest task success among memory-augmented methods, reinforcing that rule-based Precision@5 is a useful diagnostic but not a sufficient retention objective.
 
\section{Discussion and Limitations}
\label{sec:discussion}
 
External memory robustly improves ALFWorld task success across two seeds and four conditions, with paired sign tests significant in every memory-vs-no-memory comparison except Unbounded and FIFO under noisy writes. Bounded retention at $K{=}50$ matches unbounded memory at $K{=}100$ on task success while halving the candidate set scored at retrieval time, with no statistically meaningful loss at our sample sizes. The noisy-write stress reveals memory pollution: unbounded memory loses task success and retrieval precision, FIFO retains task success but its Precision@5 collapses, and TraceRetain-CEM remains stable on both. Bounded retention transfers to held-out in-distribution tasks at parity with unbounded memory.
 
At $K{=}50$ with top-5 retrieval and $T{=}200$ episodes, the bounded agent scores at most 50 candidates per retrieval rather than up to 200, and the prompt token footprint per retrieval is bounded regardless of stream length. Bounded retention also reduces average environment steps by 37\% to 55\% over no memory across conditions (e.g., 18.35 to 11.61 under noisy-write, 20.98 to 9.52 on eval-seen), translating directly into fewer LLM calls per task. On clean saturated benchmarks the memory-efficiency gain comes at no measurable task-success cost, which makes the negative result useful: a deployment can adopt the simplest cache heuristic that fits its workload, and only pay the engineering cost of learned retention when streams are demonstrably noisy.
 
Clean ALFWorld at $T{=}100$ or $T{=}200$ with \texttt{gpt-5-mini} does not naturally exhibit memory pollution. Learned retention does not robustly dominate cache heuristics on clean saturated runs. TraceRetain does not consistently improve rule-based Precision@5 in the clean regime. TraceRetain-CEM is not reinforcement learning. An exploratory offline policy-gradient variant (TraceRetain-RL-K50) underperformed both TraceRetain-Linear and TraceRetain-CEM on clean $T{=}100$ seed 42 (95/100, Appendix), suggesting that offline policy gradient over interpretable features is not the right inductive bias at this scale.
 
The $T{=}200$, noisy-write, and eval-seen runs are single-seed; differences within Wilson CIs should not be over-interpreted. The noisy-write distractor design directly stresses features the CEM scorer was selected against, so the result demonstrates discrimination in a regime constructed to favor the method, not robustness to arbitrary noise. The CEM tuning subset overlaps the seed-42 train stream, so seed-43 and eval-seen are the unbiased checks. Precision@5 is rule-based and may underestimate utility, and the strong base LLM compresses dynamic range on clean tasks and likely understates retention's value at weaker base scales.
 
\newpage

\section*{Impact Statement}
 
This work studies bounded memory retention as an inference-time efficiency lever for long-horizon LLM agents. Our defensible practical takeaway---that on saturated clean benchmarks the simplest cache heuristic that fits the workload is sufficient, and that learned retention's value is contingent on demonstrably noisy streams---supports cost-aware deployment: capping the candidate set scored at retrieval and the prompt tokens consumed per episode bounds the compute and energy cost of long-running agents at no measurable task-success cost in our study. Our experimental footprint is approximately 4000 ALFWorld episodes on a frozen base model with no parameter updates at any point, modest by current standards. Beyond the standard concerns associated with deploying LLM agents---reliability, hallucinated memory contents, and the need for operator oversight when memories are written and retrieved automatically---we do not foresee specific ethical risks. We caution that our results characterize a single benchmark with a single strong base model and may not generalize to weaker agents or naturally noisier streams.

\bibliographystyle{icml2024}

\appendix

\section{Full Clean $T{=}100$ Table (Seed 42)}
\label{app:full-clean}

\begin{table}[h]
\centering
\footnotesize
\caption{Full clean $T{=}100$ baseline table at seed 42. Bracketed values are Wilson 95\% CIs. TR- prefixes denote TraceRetain variants.}
\label{tab:full-clean}
\setlength{\tabcolsep}{4pt}
\begin{tabular}{lcccc}
\toprule
Policy & Mem & Success [95\% CI] & Steps & P@5 \\
\midrule
TR-Linear-K50 & 50  & 99/100\;[95,100] & 10.44 & 16.1\% \\
Unbounded          & 100 & 99/100\;[95,100] & 10.82 & 20.2\% \\
FIFO-K50           & 50  & 97/100\;[92,99] & 11.65 & 15.8\% \\
TR-CEM-K50    & 50  & 97/100\;[92,99] & 12.14 & 16.9\% \\
LRU-K50            & 50  & 96/100\;[90,98] & 12.32 & 14.6\% \\
Random-K50         & 50  & 96/100\;[90,98] & 12.44 & 17.7\% \\
Ebbinghaus-K50     & 50  & 96/100\;[90,98] & 12.71 & 15.4\% \\
TR-RL-K50     & 50  & 95/100\;[89,98] & 12.20 & 17.7\% \\
LFU-K50            & 50  & 94/100\;[88,97] & 12.22 & 17.1\% \\
No memory          & 0   & 88/100\;[80,93] & 18.47 & 0.0\%  \\
\bottomrule
\end{tabular}
\end{table}

\section{Precision@5 Discussion}
\label{app:precision}

A retrieved memory is counted as relevant for Precision@5 if it satisfies two conditions. First, the memory must be successful, i.e., its stored trajectory completed the task. Second, either its normalized ALFWorld task key must exactly match the current task key, or it must share the current task family and match either the target object or the receptacle.

In implementation, ALFWorld task keys are normalized by removing the trailing instance identifier from the task directory name. For exposition, we write a normalized task signature as
\[
    (\text{family}, \text{object}, \text{modifier}, \text{receptacle}).
\]
Exact-key relevance uses the full normalized signature; partial relevance requires the same family and either the same object or the same receptacle.

\paragraph{Worked example.}
Suppose the current task has signature
\[
    (\text{clean-place}, \text{apple}, \text{none}, \text{fridge}).
\]
A successful prior memory with the same signature counts as relevant. A successful memory with signature
\[
    (\text{clean-place}, \text{tomato}, \text{none}, \text{fridge})
\]
also counts, because it has the same family and receptacle. A successful memory with signature
\[
    (\text{heat-place}, \text{apple}, \text{none}, \text{microwave})
\]
does not count, because the task family differs. Finally, a failed memory with the same signature as the current task does not count, because relevance requires a successful stored trajectory.

Precision@5 is the fraction of the retrieved memories, up to five, that satisfy these conditions. If fewer than five memories are available, the denominator is the number retrieved; if no memories are retrieved, precision is zero. We treat Precision@5 as a diagnostic proxy for retrieval quality rather than a primary outcome metric.

\section{Task and Memory Example}
\label{app:task-memory-example}

ALFWorld tasks are text-based household manipulation problems. A task may
ask the agent to put a clean apple in the fridge. The agent receives a
natural-language task description and interacts with the environment
through text actions such as taking an object, moving between receptacles,
cleaning the object, and placing it in the target receptacle.

After an episode, the agent writes a memory item $m_i$ summarizing the
attempt. For example, a successful memory might contain the task
description ``put a clean apple in the fridge,'' outcome = success,
environment step count, a compressed trajectory summary such as ``take
apple; clean apple in sink; go to fridge; put apple in fridge,'' the task
embedding, the creation episode, and retrieval metadata such as access
count and last retrieval time.

TraceRetain does not modify the base LLM. It only decides whether this
memory should remain in the external memory bank when capacity is reached.
A memory like the one above receives high retention score if it is
successful, specific, useful for later retrieved tasks, and not redundant
with many other stored memories. A failed or redundant memory receives a
lower score and is more likely to be evicted.

\end{document}